\RequirePackage{fix-cm}
\documentclass[twocolumn]{svjour3}          
\smartqed  
\usepackage{graphicx}
\usepackage{times}
\usepackage{epsfig}
\usepackage{epstopdf}
\usepackage{amsmath}
\usepackage{amssymb}
\usepackage[labelfont=bf]{caption}
\usepackage{tabularx}
\usepackage[table]{xcolor}
\usepackage{algorithm}
\usepackage[noend]{algpseudocode}

\usepackage[section]{placeins}
\usepackage{booktabs}
\usepackage{makecell}
\usepackage{multirow}
\usepackage{mmstyles}
\usepackage{siunitx}
\usepackage{marvosym}
\usepackage{natbib}
\definecolor{citecolor}{HTML}{0071bc}
\definecolor{tabhighlight}{HTML}{e5e5e5}
\usepackage[colorlinks,citecolor=citecolor]{hyperref}
\usepackage{bbm}
\usepackage{pifont}
\usepackage{subcaption}
\usepackage{float}

\definecolor{ForestGreen}{rgb}{0.13, 0.55, 0.13}
\definecolor{Green}{rgb}{0.0, 0.5, 0.0}
\definecolor{green(munsell)}{rgb}{0.0, 0.66, 0.47}
\definecolor{green(ryb)}{rgb}{0.4, 0.69, 0.2}
\definecolor{green(pigment)}{rgb}{0.0, 0.65, 0.31}
\definecolor{GrayXMark}{gray}{0.6}

\graphicspath{{figures/}}
\DeclareGraphicsExtensions{.pdf}
\makeatletter
\def\BState{\State\hskip-\ALG@thistlm}
\makeatother
\definecolor{myGreen}{HTML}{33FF00}
\definecolor{myRed}{HTML}{FF3030}
\definecolor{myGrey}{HTML}{AA5555}
\definecolor{myWhite}{HTML}{FFFFFF}
\definecolor{maroon}{cmyk}{0,0.87,0.68,0.32}
\definecolor{petr}{HTML}{5555FF}
\definecolor{josef}{HTML}{FF3030}

\journalname{Noname}
\begin{document}
\begin{sloppypar}

\title{MosaicFusion: Diffusion Models as Data Augmenters for Large Vocabulary Instance Segmentation}


\author{Jiahao Xie      \and
        Wei Li          \and
        Xiangtai Li     \and
        Ziwei Liu       \and
        Yew Soon Ong    \and
        Chen Change Loy
}


\institute{Jiahao Xie \at
              S-Lab, Nanyang Technological University, Singapore \\
              \email{jiahao003@ntu.edu.sg}
           \and
           Wei Li \at
              S-Lab, Nanyang Technological University, Singapore \\
              \email{wei.l@ntu.edu.sg}
            \and
           Xiangtai Li \at
              S-Lab, Nanyang Technological University, Singapore \\
              \email{xiangtai.li@ntu.edu.sg}
           \and
           Ziwei Liu \at
              S-Lab, Nanyang Technological University, Singapore \\
              \email{ziwei.liu@ntu.edu.sg}
           \and
           Yew Soon Ong \at
              Nanyang Technological University, Singapore \\
              \email{asysong@ntu.edu.sg}
           \and
           Chen Change Loy \at
              S-Lab, Nanyang Technological University, Singapore \\
              \email{ccloy@ntu.edu.sg}
}
\date{Received: date / Accepted: date}

\maketitle

\begin{abstract}
We present MosaicFusion, a simple yet effective diffusion-based data augmentation approach for large vocabulary instance segmentation. Our method is training-free and does not rely on any label supervision. Two key designs enable us to employ an off-the-shelf text-to-image diffusion model as a useful dataset generator for object instances and mask annotations. First, we divide an image canvas into several regions and perform a single round of diffusion process to generate multiple instances simultaneously, conditioning on different text prompts. Second, we obtain corresponding instance masks by aggregating cross-attention maps associated with object prompts across layers and diffusion time steps, followed by simple thresholding and edge-aware refinement processing. Without bells and whistles, our MosaicFusion can produce a significant amount of synthetic labeled data for both rare and novel categories.
Experimental results on the challenging LVIS long-tailed and open-vocabulary benchmarks demonstrate that MosaicFusion can significantly improve the performance of existing instance segmentation models, especially for rare and novel categories.
Code: \url{https://github.com/Jiahao000/MosaicFusion}.

\end{abstract}

\section{Introduction}
\label{sec:intro}

\begin{figure*}[t]
    \centering 
    \includegraphics[width=\linewidth]{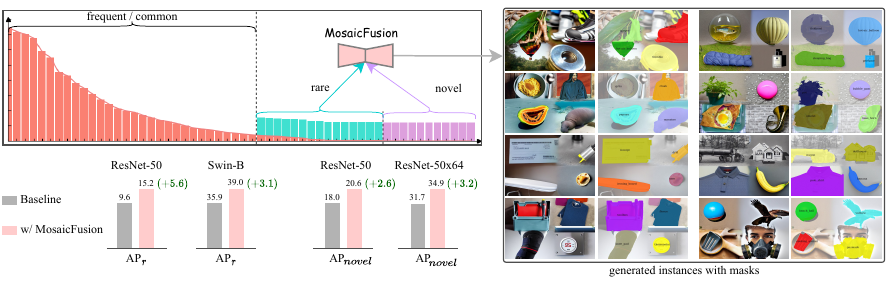}
    \caption{\textbf{Long-tailed and open-vocabulary instance segmentation} on LVIS~\citep{gupta2019lvis} using our MosaicFusion data augmentation approach, which can generate meaningful synthetic labeled data for both rare and novel categories \emph{without} further training and label supervision. We evaluate the model with the standard mask AP (\ie, AP$_{\text{r}}$ and AP$_{\text{novel}}$). MosaicFusion provides strong gains on all considered baseline methods (\eg, Mask R-CNN~\citep{he2017mask} with ResNet-50, Box-Supervised CenterNet2~\citep{detic} with Swin-B, F-VLM~\citep{kuo2023f} with ResNet-50 and ResNet-50x64)}
    \label{fig:teaser}
\end{figure*}

Instance segmentation is a fundamental yet challenging task---identifying and segmenting each object in an image---that empowers remarkable applications in autonomous driving, robotics and medical imaging~\citep{lin2014microsoft,cordts2016cityscapes,gupta2019lvis,waqas2019isaid,kuznetsova2020open}. 
However, manually labeling a large-scale instance segmentation dataset is extremely laborious and expensive as annotators need to provide a mask with precise boundaries and assign a unique label for each object instance. The cost of such a dense labeling process increases dramatically for a very large vocabulary, due to highly diverse and complicated visual scenes in different applications. 
As a result, it is prohibitive to scale up the vocabulary size of instance segmentation datasets.
As illustrated in Fig.~\ref{fig:teaser}, such a predicament of data scarcity becomes even worse under a natural data distribution that contains low-sample rare categories and out-of-distributed novel categories, both of which lead to poor performance of state-of-the-art instance segmentation models in long-tailed and open-vocabulary scenarios.

In the literature, researchers have sought ways to reduce the heavy reliance on labeled training data for instance segmentation. One popular method, Copy-Paste~\citep{ghiasi2021simple}, enhances datasets by placing objects onto backgrounds. However, while effective, this straightforward technique may not offer the extensive diversity and high-quality masks needed for optimal model training.
Recently, a promising line of research~\citep{zhang2021datasetgan,baranchuk2021label,li2022bigdatasetgan} builds upon deep generative models to synthesize the massive number of images with pixel-level labels. However, existing approaches rely on training auxiliary segmentation architectures on top of GANs~\citep{zhang2021datasetgan,li2022bigdatasetgan} or diffusion models~\citep{baranchuk2021label}, limiting their applications only to a pre-defined set of classes.

In this study, we aim to address the challenge of limited labeled training data for large vocabulary instance segmentation, especially in long-tailed and open-set scenarios.
Driven by the groundbreaking advancements in large-scale text-to-image (T2I) diffusion models---like Google's Imagen~\citep{saharia2022photorealistic}, OpenAI's DALL-E 2~\citep{ramesh2022hierarchical}, and Stability AI's Stable Diffusion~\citep{rombach2022high}---which excel in generating photorealistic images from free-form text prompts, we delve into producing vast amounts of synthetic data using T2I models to enhance instance segmentation training.
Yet, this promising avenue comes with two primary hurdles: 1) How can we produce high-quality scene images featuring multiple objects? 2) How can we derive the corresponding per-pixel instance labels, namely, instance masks, from these diffusion models without additional model training or label supervision?

To this end, we propose MosaicFusion, a diffusion-based data augmentation pipeline to generate images and masks simultaneously for large vocabulary instance segmentation.
Our method consists of two components: image generation and mask generation. For image generation, we first divide an image canvas into several regions. We then run the diffusion process on each region simultaneously, using the shared noise prediction model, while conditioning on a different text prompt. 
In this way, we can control the diffusion model to generate multiple objects at specific locations within a single image.
For mask generation, we first aggregate the cross-attention maps of the text token corresponding to a certain object across different layers and time steps in the diffusion process. We then threshold the aggregated attention maps and use standard edge-aware refinement algorithms, such as Bilateral Solver (BS)~\citep{barron2016fast}, to further refine the mask boundaries. The generated images and corresponding instance masks are finally used as a synthetic labeled dataset to train off-the-shelf models for instance segmentation.

Overall, our main contributions are summarized as follows:

\textbf{1)} We propose MosaicFusion, an automatic diffusion-based data augmentation pipeline to expand the existing instance segmentation dataset. Our method can generate images and masks simultaneously \emph{without} relying on additional off-the-shelf object detection and segmentation models to label the data further.

\textbf{2)} Our method allows us to generate customized objects at specific locations in a single image. We study both single-object and multi-object image generation scenarios and reveal that generating images with multiple objects is more beneficial than generating those with a single object.

\textbf{3)} Extensive experiments on two challenging benchmarks, \ie, long-tailed and open-vocabulary instance segmentation on LVIS~\citep{gupta2019lvis}, demonstrate that our method can significantly improve the performance of existing object detectors and instance segmentors, especially for rare and unseen categories. Figure~\ref{fig:teaser} shows the non-trivial performance improvement achieved with MosaicFusion.

\begin{table*}[t]
\centering
\small
\caption{\textbf{Comparison with other diffusion-based data augmentation works} in terms of key properties. Our MosaicFusion is the \emph{only} method with all these desired properties}
\label{tab:concurrent}
\begin{tabular}{lcccc}
\toprule
Properties        & \cite{ge2022dall}  & \cite{zhao2023x} & \cite{li2023guiding} & Ours\\ \midrule
Training-free                          & \ding{51}   & \ding{51} &  \textcolor{GrayXMark}{\ding{55}}  & \ding{51}\\
Directly generate multiple objects     & \textcolor{GrayXMark}{\ding{55}} & \textcolor{GrayXMark}{\ding{55}} &  \textcolor{GrayXMark}{\ding{55}}  & \ding{51}\\
Agnostic to detection architectures    & \ding{51} & \ding{51} &  \textcolor{GrayXMark}{\ding{55}}  & \ding{51}\\
Without extra detectors or segmentors  & \textcolor{GrayXMark}{\ding{55}}  & \textcolor{GrayXMark}{\ding{55}} &  \textcolor{GrayXMark}{\ding{55}}  & \ding{51}\\ \bottomrule
\end{tabular}
\end{table*}

\section{Related Work} 
\label{sec:work}

\noindent\textbf{Text-to-image (T2I) diffusion models.}
Recent advances in large-scale generative models, such as Imagen~\citep{saharia2022photorealistic}, DALL-E 2~\citep{ramesh2022hierarchical}, and Stable Diffusion~\citep{rombach2022high}, have brought significant progress in AI-powered image creation by training on internet-scale text-image datasets. These models can be conditioned on free-form text prompts to generate photorealistic images. This enables improved controllability in personalized image generation~\citep{gal2023designing}, content editing~\citep{hertz2023prompt}, zero-shot  translation~\citep{parmar2023zero}, and concept customization~\citep{kumari2022multi}.
Such great flexibility and scalability also bring the potential of transforming a T2I model as an effective training data generator.
In particular, our work is more related to a strand of recent research~\citep{hertz2023prompt,chefer2023attend,xie2023boxdiff,phung2023grounded} that uses cross-attention maps in T2I models for image synthesis and editing. Despite the great success, it remains unclear to what extent the vision-language correspondence residing in cross-attention maps can benefit visual perception tasks like instance segmentation. As opposed to their works, we make the first attempt to leverage such a \emph{free} correspondence in diffusion models to synthesize images and masks for large vocabulary instance segmentation tasks.

\noindent\textbf{Data augmentation for instance segmentation.}
Instance segmentation models are data-hungry and label-expensive. Therefore, many works aim at improving the performance from the data augmentation perspective. Several earlier works adopt synthesis methods via rendering graphics~\citep{su2015render,hinterstoisser2018pre} or copying from computer games~\citep{richter2016playing}. Due to the huge domain gap between synthetic and real data, another line of works~\citep{dwibedi2017cut,dvornik2018modeling,fang2019instaboost,xie2021unsupervised,ghiasi2021simple} use real image sets~\citep{gupta2019lvis}. For instance, Copy-Paste~\citep{ghiasi2021simple} shows that pasting objects onto the background using their masks can work well. However, these methods are not scalable for large vocabulary settings since the augmented instances are still confined to existing ones in the training data, \ie, they cannot create new instances for rare or novel categories with substantially more diversity.
In contrast, our goal is to generate multiple diverse rare or novel objects on the same image with their masks. Our method is orthogonal to prior works using real data for augmentation, as verified in Sect.~\ref{subsec:properties}. Concurrently, several works~\citep{ge2022dall,zhao2023x,li2023guiding} also use diffusion models for instance segmentation augmentation. The comparisons with these works are summarized in Table~\ref{tab:concurrent}. Our MosaicFusion is the \emph{only} method that is training-free, able to directly generate multiple objects and corresponding masks without relying on off-the-shelf detectors or segmentors, and compatible with various detection architectures.

\noindent\textbf{Long-tailed instance segmentation.}
This task aims to handle class imbalance problems in instance segmentation. Most approaches adopt data re-sampling~\citep{gupta2019lvis,liu2020deep,wu2020forest}, loss re-weighting~\citep{ren2020balanced,tan2020equalizationv1,tan2021equalizationv2,zhang2021distribution,wang2021adaptive} and decoupled training~\citep{li2020overcoming,wang2020devil}. In particular, several studies~\citep{liu2020deep,zhan2020online,xie2022delving} adopt image-level re-sampling. However, these approaches result in bias of instance co-occurrence. To deal with this issue, several works~\citep{hu2020learning,wu2020forest,zang2021fasa} perform more fine-grained re-sampling at the instance or feature level. For loss re-weighting, most approaches~\citep{ren2020balanced,tan2020equalizationv1,wang2021seesaw} rebalance the ratio of positive and negative samples during training. Meanwhile, decoupled training methods~\citep{li2020overcoming,wang2020devil} introduce different calibration frameworks to improve classification results. In contrast, MosaicFusion focuses on data augmentation and improving different detectors by generating new rare class examples.

\noindent\textbf{Open-vocabulary detection and segmentation.}
This task aims to detect and segment novel categories in a large concept space with the help of pre-trained vision-language models (VLMs)~\citep{zhang2023vision,wu2023towards}. OVR-CNN~\citep{zareian2021open} first puts forth the concept of open-vocabulary object detection. It is pre-trained on image-caption data to recognize novel objects and then fine-tuned for zero-shot detection. With the development of VLMs~\citep{CLIP,ALIGN}, ViLD~\citep{ViLD} is the first work to distill the rich representations of pre-trained CLIP~\citep{CLIP} into the detector. Subsequently, many studies~\citep{LSeg,zhong2022regionclip,OV-DETR,OpenSeg,du2022learning,gao2022open,minderer2022simple,chen2022open,rasheed2022bridging,kuo2023f,xu2023dst,zang2024contextual} propose different ways to adapt VLM knowledge into open-vocabulary detection and segmentation.
For example, DetPro~\citep{du2022learning} introduces a fine-grained automatic prompt learning scheme, and F-VLM~\citep{kuo2023f} adopts frozen VLMs to output novel categories from cropped CLIP features directly.
Another related work, Detic~\citep{detic}, improves the performance on novel categories with the extra large-scale image classification dataset (\ie, ImageNet-21K~\citep{deng2009imagenet}) by supervising the max-size proposal with all image labels. However, it needs more training data and the vocabulary size is limited by the classification dataset. As verified in Sect.~\ref{subsec:comparison}, MosaicFusion is orthogonal to the CLIP knowledge in the open-vocabulary setting, which boosts the state-of-the-art F-VLM~\citep{kuo2023f} by a significant margin.

\section{MosaicFusion}
\label{sec:method}

\begin{figure*}[t]
	\centering
	\includegraphics[width=\linewidth]{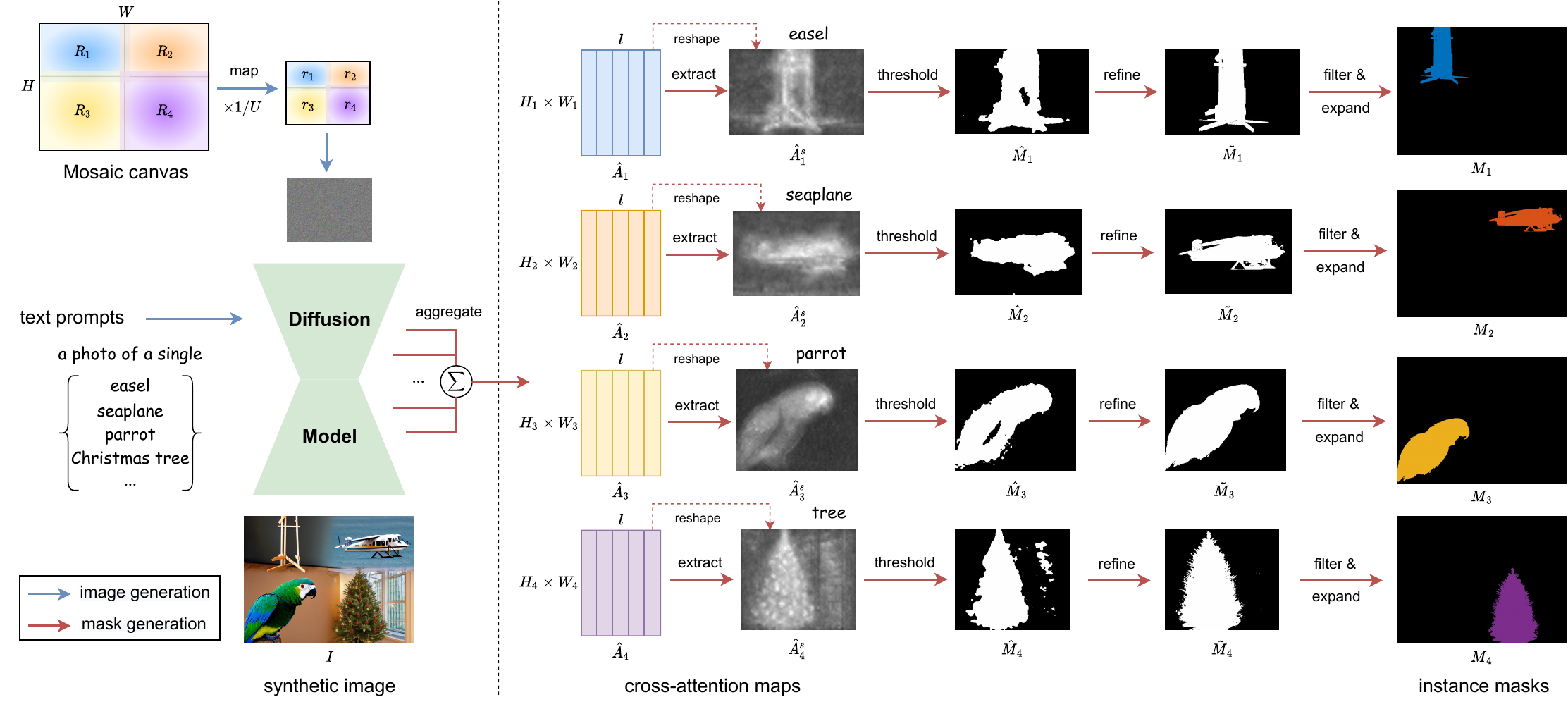}
	\caption{\textbf{Overview of our MosaicFusion pipeline.} The \emph{left} part shows the image generation process, while the \emph{right} part shows the mask generation process. Given a user-defined Mosaic image canvas and a set of text prompts, we first map the image canvas from the pixel space into the latent space. We then run the diffusion process on each latent region parallelly with the shared noise prediction model, starting from the same initialization noise while conditioning on different text prompts, to generate the synthetic image with multiple objects specified in each region. Simultaneously, we aggregate the region-wise cross-attention maps for each subject token by upscaling them to the original region size in the pixel space and averaging them across all attention heads, layers, and time steps. After that, we binarize the aggregated attention maps, refine the boundaries, filter out the low-quality masks, and expand them to the size of the whole image canvas to obtain the final instance masks}
	\label{fig:pipeline}
\end{figure*}

Our MosaicFusion is a \textit{training-free} diffusion-based dataset augmentation pipeline that can produce image and mask pairs with multiple objects simultaneously using the off-the-shelf text-to-image diffusion models. The overall pipeline of our approach is illustrated in Fig.~\ref{fig:pipeline}. It contains two components: image generation and mask generation. In the following subsections, we first introduce some preliminaries \wrt latent diffusion models and cross-attention layers in Sect.~\ref{subsec:preliminary}. We then detail the image and mask generation process of MosaicFusion in Sect.~\ref{subsec:image} and Sect.~\ref{subsec:mask}, respectively.

\subsection{Preliminary}
\label{subsec:preliminary}

\noindent\textbf{Stable Diffusion.}
We build our data generation framework upon the state-of-the-art text-to-image latent diffusion model, \ie, Stable Diffusion (SD)~\citep{rombach2022high}. SD runs the diffusion process in a compressed latent space rather than the pixel space for efficiency. It consists of three components: \romannumeral1) a variational autoencoder (VAE)~\citep{kingma2014auto} that encodes and decodes the latent vectors for images; \romannumeral2) a time-conditional U-Net~\citep{ronneberger2015u} that denoises the latent vectors at varying time steps; \romannumeral3) a text encoder like CLIP~\citep{radford2021learning} that encodes the text inputs into textural embeddings. The pre-trained VAE encodes images as latent vectors for diffusion training. During inference, the generation process starts from a random Gaussian noise $z^T\sim\mathcal{N}(0,1)$ in the latent space, and is iteratively denoised conditioned on the text prompt $p$ that is encoded via the text encoder. Specifically, at each denoising step $t=1,...,T$, given $z^t$ and $p$, U-Net predicts the noise estimation term $\epsilon$ and subtracts it from $z^t$ to obtain $z^{t-1}$. The final denoised latent $z^0$ is then passed to the VAE decoder $\mathcal{D}$ to obtain the generated image $I=\mathcal{D}(z^0)$.

\noindent\textbf{Cross-attention layers.}
The vision-language interaction occurs during the denoising process in the U-Net backbone, where the embeddings of visual and textual features are fused using cross-attention layers that produce spatial attention maps for each textual token. Specifically, at each time step $t$, the spatial feature map of the noisy image $z^t$ is linearly projected into queries and reshaped as $Q^t\in\mathbb{R}^{n\times hw\times d}$. Similarly, the text prompt $p$ is first encoded into textual embeddings via a text encoder, and then linearly projected into keys $K\in\mathbb{R}^{n\times l\times d}$ and values $V\in\mathbb{R}^{n\times l\times d}$, respectively. The attention maps are the product between queries and keys:
\begin{equation}
A^t=\text{Softmax}\left(\frac{Q^tK^\mathsf{T}}{\sqrt{d}}\right), A^t\in\mathbb{R}^{n\times hw\times l}
\end{equation}
where $d$ is the latent projection dimension, $n$ is the number of attention heads, $h$ and $w$ are the height and width of the spatial feature map of the noisy image, and $l$ is the number of text tokens in the text prompt. Intuitively, $A^t[:,i,j]$ defines the probability assigned to the token $j$ for the pixel $i$ of the spatial feature map. Therefore, they can be used as a potential source to generate segmentation masks for specific text tokens.

\subsection{Image Generation}
\label{subsec:image}

\noindent\textbf{Mosaic canvas.}
To generate multiple objects at specific locations in a single image, we first need to define an image canvas to customize a region of interest for each object. Formally, given a $H\times W$ image canvas, we divide it into a set of regions $\mathcal{R}=\{R_i=(X_i,Y_i,W_i,H_i),i=1,...,N\}$ in a Mosaic style. Here, without loss of generality, we take $N=4$ as an example. Specifically, we first randomly jitter the center of the image canvas within a certain ratio range of $W$ and $H$ to obtain a Mosaic center $(x,y),x\in[\sigma W,(1-\sigma)W],y\in[\sigma H,(1-\sigma)H]$, where $\sigma\in(0,0.5]$ is a pre-defined ratio. We then use the Mosaic center $(x,y)$ as the intersections of four candidate regions, \ie, $R_1=(0,0,x,y),R_2=(x,0,W-x,y),R_3=(0,y,x,H-y),R_4=(x,y,W-x,H-y)$. To make different object regions more smoothly transitioned, we further allow a certain overlap for each neighborhood region. Let $\delta_x$, $\delta_y$ denote the number of overlapped pixels in the horizontal and vertical direction of the canvas, respectively, we can thus obtain the final region coordinates: $R_1=(0,0,x+\delta_x/2,y+\delta_y/2),R_2=(x-\delta_x/2,0,W-x+\delta_x/2,y+\delta_y/2),R_3=(0,y-\delta_y/2,x+\delta_x/2,H-y+\delta_y/2),R_4=(x-\delta_x/2,y-\delta_y/2,W-x+\delta_x/2,H-y+\delta_y/2)$.

\noindent\textbf{Prompt template.}
Given a Mosaic image canvas $\mathcal{R}$, we then consider what kind of object, \ie, class of interest, to generate for each region. This is achieved by defining a customized text prompt for each region. Specifically, we use a generic text prompt template like ``a photo of a single\footnote{We use the word ``single'' to encourage the diffusion model to generate a single object at each specific location.} $c$, $d_c$'', where $c$ is the category name, $d_c$ is the category definition\footnote{We empirically find that appending a category definition after the category name reduces the semantic ambiguity of generated images due to the polysemy of some category names. For LVIS~\citep{gupta2019lvis}, the category definitions are readily available in annotations, where the meanings are mostly derived from WordNet~\citep{miller1995wordnet}. See Table~\ref{tab:ablation-lt-prompt} for ablations on the effect of different prompt templates.}. We randomly select $N$ category names $c$ associated with $d_c$ from a pre-defined set of interest categories to obtain $N$ text prompts $\{p_i,...,p_N\}$. We then assign prompt $p_i$ to region $R_i$ for the subsequent diffusion process, described next.

\noindent\textbf{Diffusion process.}
Since latent diffusion models like SD run the diffusion process in the latent space, we need to adapt the image canvas from the pixel space into the latent space. Due to the fully convolutional nature of U-Net (and VAE), the coordinates in the latent space can be simply mapped to the pixel space by multiplying an upscaling factor $U$, whose value is equal to the upscaling factor of U-Net in SD, \ie, $U=8$. Thus, given the region coordinates $R_i=(X_i,Y_i,W_i,H_i)$ in the pixel space, we can obtain the corresponding region coordinates $r_i=(x_i,y_i,w_i,h_i)=(X_i/U,Y_i/U,W_i/U,H_i/U)$ in the latent space. To avoid fractional indices, we ensure $(X_i,Y_i,W_i,H_i)$ in the pixel space are divisible by $U$. After mapping the image canvas into the latent canvas, we then run $N$ diffusion processes on different latent regions $r_i$ simultaneously with a shared noise-prediction network, starting from the same initialization noise while conditioning on different text prompts $p_i$. Since all diffusion processes share the same network, the memory cost is essentially the same as that of the largest region affected by a single diffusion process. As a result, we can obtain a generated image with multiple objects specified in each region.

\subsection{Mask Generation}
\label{subsec:mask}

\noindent\textbf{Attention aggregation.}
As introduced in Sect.~\ref{subsec:preliminary}, the cross-attention maps play an important role in passing the information from text tokens to spatial image features. They indicate the influence of each token on each pixel and potentially serve as a good mask source. Formally, given the noised region latent $z_i^t$ and the corresponding text prompt $p_i,i=1,...,N$, we can obtain the cross-attention maps $A_i^t$ through the U-Net network during the diffusion process formulated in Sect.~\ref{subsec:image}. Since U-Net contains a series of downsampling and upsampling blocks, we can obtain cross-attention maps with varying scales from different layers. To make full use of the spatial and temporal information, we first bicubically resize all attention maps to the original region size $(H_i,W_i)$ in the pixel space, and then average them across all attention heads, layers and time steps\footnote{We show more details in the experiment section (see Table~\ref{tab:ablation-lt-layer}, Table~\ref{tab:ablation-lt-time}, and Fig.~\ref{fig:supp_atten}) that averaging cross-attention maps across all layers and time steps is necessary to achieve the best performance.}, producing the final aggregated attention maps $\hat{A}_i\in\mathbb{R}^{H_i\times W_i\times l}$. We extract the specific attention map $\hat{A}_i^s\in\mathbb{R}^{H_i\times W_i}$ along the last channel dimension of $\hat{A}_i$ for the subject token $s$ that contains the interest category name $c$. We then normalize $\hat{A}_i^s$ within $[0,1]$ and threshold it to a binary region mask $\hat{M}_{i}\in\{0,1\}^{H_i\times W_i}$. In practice, we use Otsu's method~\citep{otsu1979threshold} to automatically determine the binary threshold.

\noindent\textbf{Edge refinement.}
The binarized attention maps provide relatively coarse masks of objects, as shown in Fig.~\ref{fig:pipeline}. To further refine the mask boundaries, we adopt the standard edge refinement post-processing techniques such as Bilateral Solver (BS)~\citep{barron2016fast} on top of the obtained coarse masks $\hat{M}_{i}$ to generate fine-grained masks $\tilde{M}_{i}\in\{0,1\}^{H_i\times W_i}$.

\noindent\textbf{Mask filtering.}
To further remove the low-quality masks, we filter the refined region masks $\tilde{M}_{i}$ based on some pre-defined criteria. Specifically, we first apply the connected component analysis~\citep{di1999simple} on the refined masks to group the pixels into several disconnected regions. We then filter out masks with areas less than 5\% or over 95\% of the whole region since these masks are highly likely segmented incorrectly. After that, we only keep region masks that have one connected component per mask since we want to generate one object at each specific location. To obtain the final instance mask $M_i\in\{0,1\}^{H\times W}$, we expand the region mask $\tilde{M}_i\in\{0,1\}^{H_i\times W_i}$ to the size of the whole image canvas by padding $0$ values for blank regions.

\section{Experiments}
\label{sec:exp}

\subsection{Implementation Details}
\label{subsec:details}

\subsubsection{Datasets}
\label{subsubsec:datasets}

We conduct our experiments of object detection and instance segmentation on the challenging LVIS v1.0 dataset~\citep{gupta2019lvis}.
LVIS is a large vocabulary instance segmentation dataset that contains 1203 categories with a long-tailed distribution of instances in each category.
It has 100k images in the training set and 19.8k images in the validation set. Based on how many images each category appears in the training set, the categories are divided into three groups: rare (1-10 images), common (11-100 images), and frequent ($>$100 images). The number of categories in each group is: rare (337), common (461), and frequent (405). In the open-vocabulary detection setting, the frequent and common categories are treated as base categories for training and the rare categories serve as novel categories for testing. The annotations of rare categories are not used during training.

\subsubsection{Evaluation Metrics}
\label{subsubsec:metrics}

We use the standard average precision (AP) as the evaluation metric, which is averaged at different IoU thresholds (from 0.5 to 0.95) across categories. We report the bounding-box AP and mask AP on all categories (denoted as AP$^{\text{box}}$, AP$^{\text{mask}}$) as well as on the rare categories (denoted as AP$^{\text{box}}_{\text{r}}$, AP$^{\text{mask}}_{\text{r}}$).
We report the average of five independent runs following the best practice of LVIS challenge~\citep{gupta2019lvis}.

\subsubsection{MosaicFusion}
\label{subsubsec:mosaicfusion}

We adopt the open-sourced Stable Diffusion v1.4 model with LMS~\citep{karras2022elucidating} scheduler. We use a guidance scale factor of 7.5 and run 50 inference steps per image for all experiments. We keep the average region size as $384\times 512$\footnote{The image resolution used during training in Stable Diffusion is $512\times 512$. We notice that the generated results will get worse if one deviates from this training resolution too much. Thus, we simply choose the aspect ratio of the average LVIS image and keep the longer dimension to 512.} to generate each object. Thus, the final image and mask size depends on the number of objects we want to generate per image, \eg, $384\times 512$ for one object, $384\times 1024$ (or $768\times 512$) for two objects, and $768\times 1024$ for four objects.

\subsubsection{Baseline Settings}
\label{subsubsec:baselines}

We consider two challenging settings to verify the effectiveness of our approach, \ie, long-tailed instance segmentation and open-vocabulary object detection. Since MosaicFusion is agnostic to the underlying detector, we consider two popular baselines in long-tailed instance segmentation: Mask R-CNN~\citep{he2017mask} and CenterNet2~\citep{zhou2021probabilistic}, and use state-of-the-art F-VLM~\citep{kuo2023f} in open-vocabulary object detection. Our implementation is based on the MMDetection~\citep{mmdetection} toolbox. We detail each baseline below.

\noindent\textbf{Mask R-CNN baseline.}
We follow the same setup in~\cite{gupta2019lvis}. Specifically, we adopt ResNet-50~\citep{he2016deep} with FPN~\citep{lin2017feature} backbone using the standard $1\times$ training schedule~\citep{mmdetection,wu2019detectron2} (90k iterations with a batch size of 16)\footnote{We are aware that different works may use different notations for a $1\times$ training schedule. In this work, we always refer $1\times$ schedule to a total of 16 $\times$ 90k images.}. We use SGD optimizer with a momentum of 0.9 and a weight decay of 0.0001. The initial learning rate is 0.02, dropped by $10\times$ at 60k and 80k iterations. Data augmentation includes horizontal flip and random resize short side $[640, 800]$, long side $<$ 1333.
We use repeat factor sampling with an oversample threshold of $10^{-3}$.

\noindent\textbf{CenterNet2 baseline.}
We follow the same setup in~\cite{detic}. Specifically, we use Swin-B~\citep{liu2021swin} with FPN backbone and a longer $4\times$ training schedule (180k iterations with a batch size of 32, \aka, Box-Supervised). We use AdamW~\citep{loshchilov2019decoupled} optimizer. The initial learning rate is 0.0001, decayed with a cosine decay schedule~\citep{loshchilov2017sgdr}. Data augmentation includes the EfficientDet~\citep{tan2020efficientdet} style large-scale jittering~\citep{ghiasi2021simple} with a training image size of $896\times 896$.
We use repeat factor sampling with an oversample threshold of $10^{-3}$ without bells and whistles.
Compared with Mask R-CNN, Box-Supervised CenterNet2 is a stronger baseline that can better verify whether MosaicFusion is scaleable with larger backbones and longer training schedules.

\noindent\textbf{F-VLM baseline.}
We follow the same setup in~\cite{kuo2023f}. Specifically, we use the pre-trained ResNet-50/ResNet-50x64 CLIP~\citep{radford2021learning} model as the frozen backbone and only train the Mask R-CNN~\citep{he2017mask} with FPN detector head for 46.1k iterations with a batch size of 256. We use SGD optimizer with a momentum of 0.9 and a weight decay of 0.0001. The initial learning rate is 0.36, dropped by $10\times$ at intervals $[0.8, 0.9, 0.95]$. Data augmentation includes large-scale jittering with a training image size of $1024\times 1024$. The base/novel VLM score weight is 0.35/0.65, respectively. We set the background weight as 0.9, and the VLM temperature as 0.01. We use CLIP prompt templates and take the average text embeddings of each category.
The state-of-the-art performance of F-VLM in open-vocabulary object detection makes it a strong baseline to verify the effectiveness of our MosaicFusion in the open-vocabulary setting.

\subsection{Main Properties}
\label{subsec:properties}

We start by ablating our MosaicFusion using Mask R-CNN R50-FPN as the default architecture on the standard LVIS long-tailed instance segmentation benchmark. Several intriguing properties are observed.

\noindent\textbf{Single object vs. multiple objects.}
We first study the effect of the number of generated objects per image. 
As shown in Table~\ref{tab:ablation-lt-object}, simply generating one ($N=1$) object per image has already outperformed the baseline by a significant margin (+4.4\% AP$_{\text{r}}$, +0.7\% AP). 
This indicates that synthetic images generated from diffusion models are indeed useful in improving the performance of instance segmentation tasks, especially for rare categories. 
Increasing the number of objects per image tends to further improve the performance, with the best performance achieved by setting $N=4$ (+5.6\% AP$_{\text{r}}$, +1.4\% AP) as more objects in a single image provide more instances and increase the task difficulty. It is worth noting that a random variant, \ie, randomly selecting the number of objects per image among $N=1,2,4$ rather than setting a fixed number achieves the sub-optimal performance compared with $N=4$. This demonstrates that the number of objects matters more than the diversity of the spatial layout of the image canvas.
See Fig.~\ref{fig:supp_vlz} for the visualization of some corresponding example images with the different number of generated objects.

\noindent\textbf{Mosaic canvas design.}
We then study the different design choices of the Mosaic image canvas in Table~\ref{tab:ablation-lt-center} and Table~\ref{tab:ablation-lt-overlap}.

Table~\ref{tab:ablation-lt-center} ablates the effect of the jittering ratio of the Mosaic center. 
A larger ratio creates sub-regions with more diverse shapes. 
MosaicFusion works the best with a moderate jittering ratio (\ie, $\sigma=0.375$). Jittering the center with a larger ratio tends to distort the image generation quality, especially for those of smaller regions.

Table~\ref{tab:ablation-lt-overlap} further ablates the effect of the number of overlapped pixels for each neighborhood region. Generating each object with a moderate region overlap (\ie, $\delta_x=64$, $\delta_y=48$) performs better than strictly separating each region for each object (\ie, $\delta_x=0$, $\delta_y=0$) since a certain overlap can allow a more smooth transition between different object regions, having a harmonization effect on the whole image. 
However, creating sub-regions with a larger overlap tends to degrade the performance as different objects will be highly entangled with each other.

\noindent\textbf{Text prompt design.}
We compare different text prompt templates in Table~\ref{tab:ablation-lt-prompt}. Simply using the category name $c$ as the text prompt has already surpassed the baseline by a clear margin (+3.9\% AP$_{\text{r}}$, +1.1\% AP). Decorating $c$ with the prefix ``a photo of a single'' slightly improves the performance upon $c$ (+0.5\% AP$_{\text{r}}$, +0.2\% AP). Appending the category definition $d_c$ after $c$ leads to further performance gains (+1.2\% AP$_{\text{r}}$, +0.1\% AP). We believe that there exist better text prompts and leave more in-depth study on prompt design for future work.

\noindent\textbf{Generated category types.}
Table~\ref{tab:ablation-lt-cateogry} studies the types of interest categories to generate. The results demonstrate that generating images with objects from all categories (\ie, rare, common, and frequent, denoted as \{r, c, f\}) leads to better performance than generating those with objects from rare categories only (denoted as \{r\}). The improvements are consistent regardless of the metrics on rare categories or all categories (+1.0\% AP$_{\text{r}}$, +0.7\% AP).

\begin{table*}[t]
\centering
\small
\caption{\textbf{Ablations for MosaicFusion on LVIS long-tailed instance segmentation.} We use Mask R-CNN R50-FPN ($1\times$ schedule). The baseline is trained on the original \emph{training} set, while others are trained on the \emph{training+synthetic} set. We report mask AP on the \emph{validation} set. Unless otherwise specified, the default settings are: (a) the number of generated objects per image is 4, (b) the center jitter ratio is 0.375, (c) the number of overlapped pixels is 64 in the horizontal direction and 48 in the vertical direction, respectively, (d) the text prompt template is ``a photo of a single $c$, $d_c$'', (e) the generated category types are all categories (\ie, rare, common, and frequent), (f) the number of generated images per category is 25, (g-h) the cross-attention maps are aggregated across all attention layers and time steps, and (i) the diffusion model is Stable Diffusion v1.4. The default entry is marked in \colorbox{gray!20}{gray}}
\label{tab:ablation-lt}
\vspace{+8pt}
\begin{minipage}[t]{.3\linewidth}
\centering
\subcaption{\textbf{Number of objects.} Increasing the number of objects per image performs better}
\label{tab:ablation-lt-object}
\begin{tabular}{ccc}
\toprule
$N$ & AP$_{\text{r}}$ & AP \\ \midrule
- & 9.6 & 21.7 \\ \midrule
1 & 14.0 & 22.4 \\
2 & 13.3 & 22.5 \\
4 & \cellcolor{gray!20}\textbf{15.2}  & \cellcolor{gray!20}\textbf{23.1}\\
random & 14.5 & 22.9 \\ \bottomrule
\end{tabular}
\end{minipage}\hfill
\begin{minipage}[t]{.3\linewidth}
\centering
\subcaption{\textbf{Center jitter ratio.} Moderately jittering the canvas center works the best}
\label{tab:ablation-lt-center}
\begin{tabular}{ccc}
\toprule
$\sigma$ & AP$_{\text{r}}$ & AP \\ \midrule
- & 9.6 & 21.7 \\ \midrule
0.25 & 13.8 & 22.9 \\
0.375 & \cellcolor{gray!20}\textbf{15.2}  & \cellcolor{gray!20}\textbf{23.1} \\
0.5 & 14.2 & 23.0 \\ \bottomrule
\end{tabular}
\end{minipage}\hfill
\begin{minipage}[t]{.3\linewidth}
\centering
\subcaption{\textbf{Overlapped pixels.} Allowing a certain region overlap in the canvas is effective}
\label{tab:ablation-lt-overlap}
\begin{tabular}{ccc}
\toprule
$(\delta_x,\delta_y)$ & AP$_{\text{r}}$ & AP \\ \midrule
- & 9.6 & 21.7 \\ \midrule
(0, 0) & 14.2 & 23.0 \\
(64, 48) & \cellcolor{gray!20}\textbf{15.2}  & \cellcolor{gray!20}\textbf{23.1} \\
(128, 96) & 13.2 & 22.9 \\ \bottomrule
\end{tabular}
\end{minipage}
\vfill
\vspace{+10pt}
\begin{minipage}[t]{.3\linewidth}
\centering
\subcaption{\textbf{Text prompt.} Appending a category definition is more accurate}
\label{tab:ablation-lt-prompt}
\begin{tabular}{ccc}
\toprule
Prompt template & AP$_{\text{r}}$ & AP \\ \midrule
- & 9.6 & 21.7 \\ \midrule
$c$ & 13.5 & 22.8 \\
a photo of a single $c$ & 14.0  & 23.0 \\
a photo of a single $c$, $d_c$ & \cellcolor{gray!20}\textbf{15.2} & \cellcolor{gray!20}\textbf{23.1} \\ \bottomrule
\end{tabular}
\end{minipage}\hfill
\begin{minipage}[t]{.3\linewidth}
\centering
\subcaption{\textbf{Category type.} Generating all categories leads to more gains}
\label{tab:ablation-lt-cateogry}
\begin{tabular}{ccc}
\toprule
Category set & AP$_{\text{r}}$ & AP \\ \midrule
- & 9.6 & 21.7 \\ \midrule
\{r\} & 14.2  & 22.4 \\
\{r, c, f\} & \cellcolor{gray!20}\textbf{15.2}  & \cellcolor{gray!20}\textbf{23.1} \\ \bottomrule
\end{tabular}
\end{minipage}\hfill
\begin{minipage}[t]{.3\linewidth}
\centering
\subcaption{\textbf{Number of images.} Generating 25 images per category is enough}
\label{tab:ablation-lt-image}
\begin{tabular}{ccc}
\toprule
\# Images & AP$_{\text{r}}$ & AP \\ \midrule
- & 9.6 & 21.7 \\ \midrule
10 & 13.1 & 22.5 \\
25 & \cellcolor{gray!20}\textbf{15.2} & \cellcolor{gray!20}\textbf{23.1} \\
50 & 14.7 & 23.0 \\ \bottomrule
\end{tabular}
\end{minipage}
\vfill
\vspace{+10pt}
\begin{minipage}[t]{.3\linewidth}
\centering
\subcaption{\textbf{Attention layer.} All layers contribute positively to the results}
\label{tab:ablation-lt-layer}
\begin{tabular}{ccc}
\toprule
Layer resolution & AP$_{\text{r}}$ & AP \\ \midrule
- & 9.6 & 21.7 \\ \midrule
$\leq\times 1/32$ & 12.6 & 22.7 \\
$\leq\times 1/16$ & 14.3 & 22.9 \\
$\leq\times 1/8$ & \cellcolor{gray!20}\textbf{15.2} & \cellcolor{gray!20}\textbf{23.1} \\ \bottomrule
\end{tabular}
\end{minipage}\hfill
\begin{minipage}[t]{.3\linewidth}
\centering
\subcaption{\textbf{Time step.} All time steps are necessary for the best performance}
\label{tab:ablation-lt-time}
\begin{tabular}{ccc}
\toprule
Time step & AP$_{\text{r}}$ & AP \\ \midrule
- & 9.6 & 21.7 \\ \midrule
$\leq$ 10 & 14.3 & 22.8 \\
$\leq$ 25 & 14.6 & 22.9 \\
$\leq$ 50 & \cellcolor{gray!20}\textbf{15.2} & \cellcolor{gray!20}\textbf{23.1} \\ \bottomrule
\end{tabular}
\end{minipage}\hfill
\begin{minipage}[t]{.3\linewidth}
\centering
\subcaption{\textbf{Diffusion model.} MosaicFusion benefits from more advanced diffusion models}
\label{tab:ablation-lt-sd}
\begin{tabular}{ccc}
\toprule
Stable Diffusion & AP$_{\text{r}}$ & AP \\ \midrule
- & 9.6 & 21.7 \\ \midrule
v1.2 & 12.7 & 22.6 \\
v1.3 & 14.4 & 22.9 \\
v1.4 & \cellcolor{gray!20}{15.2} & \cellcolor{gray!20}{23.1} \\
v1.5 & \textbf{15.5}  & \textbf{23.2} \\ \bottomrule
\end{tabular}
\end{minipage}
\vspace{+8pt}
\end{table*}

\noindent\textbf{Generated image numbers.}
Table~\ref{tab:ablation-lt-image} studies the effect of the number of generated images per interest category\footnote{Note that the final number of synthetic images per category is not fixed as we need to filter some images during the mask filtering process introduced in Sect.~\ref{subsec:mask}. We observe that $\sim$50\% generated images and masks will be discarded accordingly.}. \emph{Only} generating 10 images per category has already improved upon the baseline by a significant margin (+3.5\% AP$_{\text{r}}$, +0.8\% AP). A better performance is achieved by setting the number of generated images per category as 25 (+5.6\% AP$_{\text{r}}$, +1.4\% AP). Generating more images does not lead to further performance gains. We hypothesize that the diversity of generated images from existing diffusion models is a main challenge to scale up. We expect further performance improvement to be achieved with the emergence of more expressive diffusion models in the future.

\noindent\textbf{Attention layers and time steps.}
We study whether aggregating cross-attention maps across all attention layers and time steps is necessary. Table~\ref{tab:ablation-lt-layer} and Table~\ref{tab:ablation-lt-time} ablate the influence of layer resolutions and time steps, respectively. The results indicate that all layers and time steps contribute positively to the performance, proving that we should make full use of the spatial and temporal information during the diffusion process to produce the highest-quality attention maps for instance segmentation.
See Fig.~\ref{fig:supp_atten} for the visualization of some example cross-attention maps across different layers and time steps.

\noindent\textbf{Diffusion models.}
In Table~\ref{tab:ablation-lt-sd}, we study whether our MosaicFusion benefits from the improvement of diffusion models. We use Stable Diffusion v1.x (\eg, v1.2, v1.3, v1.4, and v1.5) as examples for ablations. A better diffusion model generally leads to better instance segmentation performance. This makes sense as a more advanced diffusion model generates higher-quality images.

\begin{table}[t]
\centering
\small
\caption{\textbf{Comparison with Mosaic~\citep{bochkovskiy2020yolov4} data augmentation.} We use Mask R-CNN R50-FPN ($1\times$ schedule) as the baseline and report mask AP. Synthesizing multi-object images is more effective than simply combining single-object images}
\label{tab:analysis-mosaic}
\vspace{+5pt}
\begin{tabular}{lcc}
\toprule
Method & AP$_{\text{r}}$ & AP\\ \midrule
Mask R-CNN~\citep{he2017mask} & 9.6 & 21.7\\
w/ MosaicFusion ($N=1$) & 14.0 & 22.4 \\
w/ MosaicFusion ($N=1$) + Mosaic aug. & 14.0 & 22.4 \\
w/ MosaicFusion ($N=4$) & \textbf{15.2} & \textbf{23.1} \\ \bottomrule
\end{tabular}
\vspace{+5pt}
\end{table}

\setcounter{table}{4}
\begin{table}[t]
\centering
\small
\caption{\textbf{Comparison between SD and SDXL~\citep{podell2023sdxl}.} We use Mask R-CNN R50-FPN ($1\times$ schedule) as the baseline and report mask AP. MosaicFusion benefits from a higher-resolution diffusion model}
\label{tab:analysis-sdxl}
\vspace{+5pt}
\resizebox{.49\textwidth}{!}{
\begin{tabular}{lccc}
\toprule
Method & Region size & AP$_{\text{r}}$ & AP\\ \midrule
Mask R-CNN~\citep{he2017mask} & - & 9.6 & 21.7\\
w/ MosaicFusion (SD v1.5) & $384\times 512$ & 15.5 & 23.2 \\
w/ MosaicFusion (SDXL base v1.0) & $768\times 1024$ & \textbf{16.2} & \textbf{23.4} \\ \bottomrule
\end{tabular}
}
\vspace{+5pt}
\end{table}

\noindent\textbf{Comparison with other data augmentation methods.}
We compare our MosaicFusion with other existing data augmentation methods for instance segmentation.

We first compare MosaicFusion with Mosaic data augmentation proposed in the popular object detection framework YOLO-v4~\citep{bochkovskiy2020yolov4}. It combines four training images into one with certain ratios after each image has been independently augmented. To ensure a fair comparison, we simply replace the multi-object images generated by MosaicFusion with the four combined generated single-object images using the Mosaic data augmentation while keeping other hyper-parameters unchanged. 
As shown in Table~\ref{tab:analysis-mosaic}, simply adopting Mosaic augmentation to produce multi-object images does not bring further gains over the single-object version of our MosaicFusion ($N=1$). In contrast, our multi-object version of MosaicFusion ($N=4$) leads to further performance improvement compared with MosaicFusion ($N=1$). This proves that using diffusion models to synthesize multi-object images is more effective than simply combining single-object images with the existing data augmentation counterpart.

We then show that our method is \emph{orthogonal} to existing data augmentation methods. Here, we use the popular Copy-Paste~\citep{ghiasi2021simple} as an example, which produces more instances by randomly copying the existing objects from one image and pasting them into another. The results are shown in Table~\ref{tab:analysis-scp}. Either Copy-Paste or MosaicFusion improves over the baseline, while MosaicFusion yields more performance gains. This makes sense as Copy-Paste only copies and pastes existing instances in the training data, whereas MosaicFusion creates new instances with substantially more diversity than existing ones. The diversity of instances matters more for rare categories. A slightly higher performance can be further achieved by combining them together, indicating that MosaicFusion is compatible with other data augmentation methods.

\setcounter{table}{3}
\begin{table}[t]
\centering
\small
\caption{\textbf{Comparison with Copy-Paste~\citep{ghiasi2021simple} data augmentation.} We use Mask R-CNN R50-FPN ($1\times$ schedule) as the baseline and report mask AP. MosaicFusion is orthogonal to existing data augmentation methods like Copy-Paste}
\label{tab:analysis-scp}
\vspace{+5pt}
\begin{tabular}{lcc}
\toprule
Method & AP$_{\text{r}}$ & AP\\ \midrule
Mask R-CNN~\citep{he2017mask} & 9.6 & 21.7\\
w/ Copy-Paste & 10.6 & 22.1 \\
w/ MosaicFusion & 15.2 & 23.1 \\
w/ MosaicFusion + Copy-Paste & \textbf{15.5} & \textbf{23.3} \\ \bottomrule
\end{tabular}
\vspace{+5pt}
\end{table}

\setcounter{table}{5}
\begin{table}[t]
\centering
\small
\caption{\textbf{Effect of extra segmentation tools.} We use Mask R-CNN R50-FPN ($1\times$ schedule) as the baseline and report mask AP. Using SAM~\citep{kirillov2023segment} as an additional mask refiner yields further gains}
\label{tab:analysis-sam}
\vspace{+5pt}
\begin{tabular}{lcc}
\toprule
Method & AP$_{\text{r}}$ & AP\\ \midrule
Mask R-CNN~\citep{he2017mask} & 9.6 & 21.7\\
w/ MosaicFusion & 15.2 & 23.1 \\
w/ MosaicFusion + SAM & \textbf{16.0} & \textbf{23.3} \\ \bottomrule
\end{tabular}
\vspace{+5pt}
\end{table}

\noindent\textbf{Effect of higher-resolution diffusion models.}
In Table~\ref{tab:ablation-lt-sd}, we have studied the effect of diffusion models and observed that a better diffusion model generally leads to better instance segmentation performance.
Here, we go beyond SD v1 and equip MosaicFusion with a more powerful high-resolution diffusion model, \eg, SDXL~\citep{podell2023sdxl}.
Specifically, we use SDXL base v1.0 to generate higher-quality images with higher resolutions. We set the average region size as $768\times 1024$ to generate each object while keeping all other settings in Table~\ref{tab:ablation-lt} unchanged. The results based on Mask R-CNN R50-FPN are shown in Table~\ref{tab:analysis-sdxl}. Using a higher-resolution diffusion model brings more gains.

\begin{table*}[t]
\centering
\small
\caption{\textbf{LVIS long-tailed instance segmentation benchmark.} We report box AP and mask AP on the \emph{validation} set. We build our method upon two representative baselines: (\romannumeral1) Mask R-CNN baseline ($1\times$ schedule) in~\cite{gupta2019lvis} using the ResNet-50 w/ FPN backbone, and (\romannumeral2) Box-Supervised CenterNet2 baseline ($4\times$ schedule) in~\cite{detic} using the Swin-B backbone. MosaicFusion improves over the baselines regardless of the architectures with increased gains for the rare categories}
\label{tab:benchmark-lt}
\addtolength{\tabcolsep}{+4pt}
\vspace{+10pt}
\begin{tabular}{llcccc}
\toprule
Method & Backbone & AP$^{\text{box}}_{\text{r}}$ & AP$^{\text{mask}}_{\text{r}}$ & AP$^{\text{box}}$ & AP$^{\text{mask}}$\\ \midrule
Mask R-CNN~\citep{he2017mask} & ResNet-50 & 9.1 & 9.6 & 22.5 & 21.7\\
w/ MosaicFusion & ResNet-50 & 14.8  & 15.2 & 24.0 & 23.1 \\
\emph{vs. baseline} &  & \textcolor{ForestGreen}{\textbf{+5.7}}  & \textcolor{ForestGreen}{\textbf{+5.6}} & \textcolor{ForestGreen}{\textbf{+1.5}} & \textcolor{ForestGreen}{\textbf{+1.4}} \\ \midrule
MosaicOS~\citep{zhang2021mosaicos} & ResNeXt-101 & - & 21.7 & - & 28.3 \\
CenterNet2~\citep{zhou2021probabilistic} & ResNeXt-101 & - & 24.6 & - & 34.9 \\
AsyncSLL~\citep{han2020joint} & ResNeSt-269 & - & 27.8 & - & 36.0 \\
SeesawLoss~\citep{wang2021seesaw} & ResNeSt-200 & - & 26.4 & - & 37.3 \\
Copy-paste~\citep{ghiasi2021simple} & EfficientNet-B7 & - & 32.1 & 41.6 & 38.1 \\
Tan \etal~\citep{tan20201st} & ResNeSt-269 & - & 28.5 & - & 38.8 \\ \midrule
Box-Supervised~\citep{detic} & Swin-B & 39.9 & 35.9 & 45.4 & 40.7 \\
w/ MosaicFusion & Swin-B & 43.3 & 39.0 & 46.1 & 41.2 \\
\emph{vs. baseline} &  & \textcolor{ForestGreen}{\textbf{+3.4}}  & \textcolor{ForestGreen}{\textbf{+3.1}} & \textcolor{ForestGreen}{\textbf{+0.7}} & \textcolor{ForestGreen}{\textbf{+0.5}} \\ \bottomrule
\end{tabular}
\vspace{+10pt}
\end{table*}

\noindent\textbf{Effect of employing extra segmentation tools.}
So far, we have shown that MosaicFusion is the first training-free method that can generate multi-object images and masks without relying on extra detection and segmentation models. Here, we discuss whether incorporating additional segmentation tools further improves the performance. We employ an interactive segmentation model, \ie, SAM~\citep{kirillov2023segment}, to further refine the generated instance masks due to its strong zero-shot generalization ability. Specifically, after obtaining the generated annotations from MosaicFusion, we prompt SAM with MosaicFusion-generated bounding boxes to produce the final instance masks. As shown in Table~\ref{tab:analysis-sam}, using SAM-refined masks leads to further gains. Thus, if the training and inference costs of extra segmentors are not a major concern, we recommend using them to further improve the performance. Nevertheless, we keep not using extra models in the whole paper to solely verify the effectiveness of using diffusion models as data augmenters.

\begin{table*}[t]
\centering
\small
\caption{\textbf{LVIS open-vocabulary instance segmentation benchmark.} We report mask AP on the \emph{validation} set. We build our method upon state-of-the-art F-VLM~\citep{kuo2023f}. Unless otherwise specified, all methods use the CLIP~\citep{radford2021learning} pre-training and fixed prompt templates. $^\ast$: our reproduced results. $^\star$: prompt optimization~\citep{zhou2022conditional} and SoCo pre-training~\citep{wei2021aligning}. $^\dag$: joint training with ImageNet-21K~\citep{deng2009imagenet}. $^\ddag$: training with CC-3M~\citep{sharma2018conceptual}}
\label{tab:benchmark-ov}
\addtolength{\tabcolsep}{+4pt}
\vspace{+10pt}
\begin{tabular}{lllcc}
\toprule
Method & Backbone & Pre-trained CLIP & AP$_{\text{novel}}$  & \textcolor{GrayXMark}{AP}\\ \midrule
ViLD~\citep{ViLD} & ResNet-50 & ViT-B/32 & 16.1 & \textcolor{GrayXMark}{22.5} \\
ViLD-Ens.~\citep{ViLD} & ResNet-50 & ViT-B/32  & 16.6 & \textcolor{GrayXMark}{25.5} \\
DetPro$^\star$~\citep{du2022learning} & ResNet-50 & ViT-B/32  & 19.8 & \textcolor{GrayXMark}{25.9} \\
Detic-ViLD$^\dag$~\citep{detic} & ResNet-50 & ViT-B/32  & 17.8 & \textcolor{GrayXMark}{26.8} \\
RegionCLIP$^\ddag$~\citep{zhong2022regionclip} & ResNet-50 & ResNet-50  & 17.1 & \textcolor{GrayXMark}{28.2} \\ \midrule
F-VLM~\citep{kuo2023f} & ResNet-50 & ResNet-50  & 18.6 & \textcolor{GrayXMark}{24.2} \\
F-VLM$^\ast$  & ResNet-50 & ResNet-50  & 18.0 & \textcolor{GrayXMark}{23.6}  \\
w/ MosaicFusion & ResNet-50 & ResNet-50  & 20.6 & \textcolor{GrayXMark}{24.0} \\
\emph{vs. baseline} &  &    & \textcolor{ForestGreen}{\textbf{+2.6}}  & \textcolor{ForestGreen}{\textbf{+0.4}} \\ \hline \midrule
ViLD~\citep{ViLD} & ResNet-152 & ViT-B/32 & 18.7 & \textcolor{GrayXMark}{23.6} \\
ViLD-Ens.~\citep{ViLD} & ResNet-152 & ViT-B/32  & 18.7 & \textcolor{GrayXMark}{26.0} \\
ViLD-Ens.~\citep{ViLD} & EfficientNet-B7 & ViT-L/14  & 21.7 & \textcolor{GrayXMark}{29.6}  \\
ViLD-Ens.~\citep{ViLD} & EfficientNet-B7 & EfficientNet-B7  & 26.3 & \textcolor{GrayXMark}{29.3} \\
DetPro-Cascade$^\star$~\citep{du2022learning} & ResNet-50 & ViT-B/32  & 20.0 & \textcolor{GrayXMark}{27.0} \\
Detic-CN2$^\dag$~\citep{detic} & ResNet-50 & ViT-B/32  & 24.6 & \textcolor{GrayXMark}{32.4} \\
MEDet$^\ddag$~\citep{chen2022open} & ResNet-50 & ViT-B/32  & 22.4 & \textcolor{GrayXMark}{34.4} \\
Centric-OVD$^\dag$~\citep{rasheed2022bridging} & ResNet-50 & ViT-B/32  & 25.2 & \textcolor{GrayXMark}{32.9} \\
RegionCLIP$^\ddag$~\citep{zhong2022regionclip} & ResNet-50x4 & ResNet-50x4  & 22.0 & \textcolor{GrayXMark}{32.3} \\
OWL-ViT~\citep{minderer2022simple} & ViT-L/14 & ViT-L/14  & 25.6 & \textcolor{GrayXMark}{34.7} \\ \midrule
F-VLM~\citep{kuo2023f} & ResNet-50x64 & ResNet-50x64  & 32.8 & \textcolor{GrayXMark}{34.9}  \\
F-VLM$^\ast$  & ResNet-50x64 & ResNet-50x64  & 31.7 & \textcolor{GrayXMark}{34.9}  \\
w/ MosaicFusion & ResNet-50x64 & ResNet-50x64  & 34.9 & \textcolor{GrayXMark}{35.3}  \\
\emph{vs. baseline} &  &    & \textcolor{ForestGreen}{\textbf{+3.2}}  & \textcolor{ForestGreen}{\textbf{+0.4}} \\ \bottomrule
\end{tabular}
\vspace{+10pt}
\end{table*}

\subsection{Comparison with Previous Methods}
\label{subsec:comparison}

We perform system-level comparisons with previous methods on the long-tailed instance segmentation benchmark as well as the open-vocabulary object detection benchmark on LVIS.

\noindent\textbf{Long-tailed instance segmentation benchmark.}
Table~\ref{tab:benchmark-lt} presents our results on the long-tailed instance segmentation benchmark. MosaicFusion yields significant performance gains over the commonly-used Mask R-CNN baseline, with the most considerable improvement coming from AP$_{\text{r}}$ (+5.7\% AP$^{\text{box}}_{\text{r}}$, +5.6\% AP$^{\text{mask}}_{\text{r}}$). We observe consistent performance improvements when building upon a stronger Box-Supervised CenterNet2 baseline, obtaining over 3\% higher AP$_{\text{r}}$ (+3.4\% AP$^{\text{box}}_{\text{r}}$, +3.1\% AP$^{\text{mask}}_{\text{r}}$). This demonstrates that MosaicFusion is agnostic to detector architectures and is an effective plug-and-play data augmentation method for existing instance segmentation models.

\noindent\textbf{Open-vocabulary object detection benchmark.}
We further demonstrate that MosaicFusion is also a promising data augmenter in the challenging open-vocabulary object detection benchmark. We build our method upon the state-of-the-art open-vocabulary detector F-VLM~\citep{kuo2023f}, which uses the pre-trained CLIP model as the frozen backbone and Mask R-CNN with FPN as the detector head. The results are shown in Table~\ref{tab:benchmark-ov}. MosaicFusion still consistently outperforms F-VLM across different backbones especially for AP$_{\text{novel}}$ (+2.6\% AP$_{\text{novel}}$ for ResNet-50, +3.2\% AP$_{\text{novel}}$ for ResNet-50x64), even though the baseline has incorporated the open-vocabulary knowledge from the CLIP model. This indicates that text-driven diffusion models and CLIP models are complementary to each other. We can achieve better performance by leveraging the benefits from both worlds.

\subsection{Further Discussion}
\label{subsec:analysis}

We further discuss how to evaluate the synthesized mask quality directly and compare MosaicFusion with other diffusion-based data augmentation methods quantitatively.
Moreover, we examine the scalability of MosaicFusion when the model performance is already saturated and the vocabulary size is extremely large.
Finally, we perform a case study to show that our general data augmentation pipeline can be potentially extended to broader multimodal understanding tasks.

\noindent\textbf{Mask quality evaluation.}
In Sect.~\ref{subsec:properties} and Sect.~\ref{subsec:comparison}, we use the standard evaluation metrics (\ie, box AP and mask AP) in instance segmentation tasks for evaluation since our goal is to improve the instance segmentation performance, which is a common evaluation protocol used in previous works. Explicitly evaluating synthetic images and masks is challenging as no ground truth is available. One na\"ive way is to manually label all synthetic images with corresponding instance masks. However, it is extremely time-consuming and expensive. Here, we provide a more cost-effective metric to directly evaluate the synthesized mask quality. Specifically, we use SAM~\citep{kirillov2023segment} as a data annotator due to its strong zero-shot generalization ability. We first prompt SAM with MosaicFusion-generated bounding boxes to produce instance masks. We then use these SAM-generated masks as ground truth and compute mean Intersection-over-Union (mIoU) between MosaicFusion-generated masks and ground truth masks. For reference, the mIoU result of MosaicFusion using the default setting in Table~\ref{tab:ablation-lt} is 77.2\%. We hope our proposed metric can inspire future works on directly evaluating synthetic images and masks for rare and novel categories.

\noindent\textbf{Comparison with other diffusion-based data augmentation methods.}
Several concurrent works~\citep{ge2022dall,zhao2023x,li2023guiding} also use diffusion models for instance segmentation augmentation. We have discussed the differences with these works in Sect.~\ref{sec:work} and Table~\ref{tab:concurrent}.
Here, we additionally provide a quantitative comparison with X-Paste~\citep{zhao2023x}\footnote{We choose X-Paste for comparison due to its open-sourced implementation. Note that \cite{li2023guiding} uses a different setting by training and testing models on its own synthetic datasets. Thus, an apple-to-apple quantitative comparison with~\cite{li2023guiding} is infeasible.}. To ensure a fair comparison, we use the same baseline as in~\cite{zhao2023x}, \ie, CenterNet2~\citep{zhou2021probabilistic} with ResNet-50 backbone for a $4\times$ training schedule (90k iterations with a batch size of 64). Considering that X-Paste uses additional retrieved images by the CLIP model to further boost its performance, we remove these retrieved images and only keep the generated ones. Table~\ref{tab:analysis-xpaste} reports the results on LVIS. Although X-Paste uses multiple extra foreground segmentation models to label the generated data further, our MosaicFusion can still achieve competitive performance with X-Paste. It should be well noted that X-Paste here uses 100k generated images while our MosaicFusion \emph{only} uses 4k generated ones to achieve such results. As shown in Table~\ref{tab:analysis-xpaste}, MosaicFusion is \emph{at least} $4.3\times$ more efficient than X-Paste. Thus, MosaicFusion is much more cost-effective than X-Paste. This demonstrates the superiority of directly synthesizing multi-object training images over pasting multiple synthesized instances onto the background to compose training images.

\begin{table}[t]
\centering
\small
\caption{\textbf{Comparison with X-Paste~\citep{zhao2023x} on LVIS.} We use CenterNet2 with ResNet-50 ($4\times$ schedule) as the baseline, and report box AP and mask AP. We re-implement both the baseline and X-Paste using the official code of X-Paste\protect\footnotemark. All entries use the same network and training settings for a fair comparison. Note that X-Paste uses 100k generated images while MosaicFusion only uses 4k generated images. The dataset synthesis cost (GPU hours) is measured on a single A100 GPU. MosaicFusion is at least $4.3\times$ more efficient than X-Paste. $^\dag$: X-Paste requires four extra segmentation models to further label the generated data, whose costs are not included here}
\label{tab:analysis-xpaste}
\vspace{+5pt}
\resizebox{.49\textwidth}{!}{
\begin{tabular}{lccccc}
\toprule
Method & GPU hours & AP$^{\text{box}}_{\text{r}}$ & AP$^{\text{mask}}_{\text{r}}$ & AP$^{\text{box}}$ & AP$^{\text{mask}}$ \\ \midrule
baseline & - & 21.0 & 19.0 & 33.9 & 30.2 \\
w/ X-Paste & 100$^\dag$ & 25.5 & 23.0 & 34.6 & 30.7 \\
w/ MosaicFusion & 23 & \textbf{25.8} & \textbf{23.4} & \textbf{34.7} & \textbf{30.9} \\ \bottomrule
\end{tabular}
}
\vspace{+5pt}
\end{table}

\footnotetext{We find that the baseline and X-Paste cannot be fully reproduced using the official code (see issues \href{https://github.com/yoctta/XPaste/issues/2}{here}). Therefore, our reproduced results are relatively lower than the original reported performance. Nevertheless, all experiments are done under the same settings for a fair comparison.}

\begin{table}[t]
\centering
\small
\caption{\textbf{Comparison with the state-of-the-art Co-DETR~\citep{zong2023detrs} detector on LVIS.} We report box AP on the \emph{validation} set}
\label{tab:analysis-codetr}
\vspace{+5pt}
\begin{tabular}{llcc}
\toprule
Method & Backbone & AP$_{\text{r}}$ & AP \\ \midrule
Co-DETR~\citep{zong2023detrs} & Swin-L & 61.4 & 64.5 \\
w/ MosaicFusion & Swin-L & 64.4 & 65.1 \\
\emph{vs. baseline} &  & \textcolor{ForestGreen}{\textbf{+3.0}} & \textcolor{ForestGreen}{\textbf{+0.6}} \\ \bottomrule
\end{tabular}
\vspace{+5pt}
\end{table}

\noindent\textbf{Scalability in performance-saturated models.}
In Sect.~\ref{subsec:comparison}, we have chosen several representative baselines to verify the effectiveness of our MosaicFusion for large vocabulary instance segmentation.
Here, we further scale up MosaicFusion from the model axis by building upon a recent state-of-the-art Co-DETR~\citep{zong2023detrs} detector to examine its effectiveness in performance-saturated models.
Specifically, we choose the Objects365~\citep{shao2019objects365} pre-trained Co-DETR with Swin-L~\citep{liu2021swin} backbone and fine-tune the detector on LVIS following the same setup in~\cite{zong2023detrs}. We fine-tune for 16 epochs with a batch size of 16. We use AdamW~\citep{loshchilov2019decoupled} optimizer with a weight decay of 0.0001. The initial learning rate is 0.0001, dropped by $10\times$ at the 8-th epoch. Data augmentation includes large-scale jittering with copy-paste~\citep{ghiasi2021simple}, with a training image size of $1536\times 1536$. The results are shown in Table~\ref{tab:analysis-codetr}.
After pre-training on Objects365 and reaching a plateau on LVIS (61.4\% AP$_{\text{r}}$), MosaicFusion still yields +3.0\% AP$_{\text{r}}$ gains over Co-DETR, demonstrating its effectiveness in scenarios where the model performance is already saturated.

\begin{table}[t]
\centering
\small
\caption{\textbf{Open-vocabulary object detection on V3Det.} We report box AP on the \emph{validation} set. We build our method upon F-VLM~\citep{kuo2023f}}
\label{tab:analysis-v3det}
\vspace{+10pt}
\begin{tabular}{llc}
\toprule
Method & Backbone & AP$_{\text{novel}}$ \\ \midrule
Detic~\citep{detic} & ResNet-50 & 6.7 \\
RegionCLIP~\citep{zhong2022regionclip} & ResNet-50 & 3.1 \\ \midrule
F-VLM~\citep{kuo2023f} & ResNet-50 & 3.9 \\
w/ MosaicFusion & ResNet-50 & 7.4 \\
\emph{vs. baseline} &  & \textcolor{ForestGreen}{\textbf{+3.5}} \\ \midrule
F-VLM~\citep{kuo2023f} & ResNet-50x64 & 7.0 \\
w/ MosaicFusion & ResNet-50x64 & 14.5 \\
\emph{vs. baseline} &  & \textcolor{ForestGreen}{\textbf{+7.5}} \\ \bottomrule
\end{tabular}
\vspace{+15pt}
\end{table}

\begin{table}[t]
\centering
\small
\caption{\textbf{Referring expression segmentation benchmark.} We report cIoU on the \emph{validation} sets of RefCOCO (RefC), RefCOCO+ (RefC+), and RefCOCOg (RefCg). We build our method upon F-LMM~\citep{wu2024f}, LISA~\citep{lai2024lisa}, and GLaMM~\citep{rasheed2024glamm}. For F-LMM, we use DeepseekVL-1.3B~\citep{lu2024deepseek} as the frozen LMM}
\label{tab:analysis-lmm}
\vspace{+10pt}
\begin{tabular}{llccc}
\toprule
Method & RefC & RefC+ & RefCg \\ \midrule
F-LMM-1.3B~\citep{wu2024f} & 75.0 & 62.8 & 68.2 \\
w/ MosaicFusion & 77.1 & 65.3 & 69.2 \\
\emph{vs. baseline} & \textcolor{ForestGreen}{\textbf{+2.1}} & \textcolor{ForestGreen}{\textbf{+2.5}} & \textcolor{ForestGreen}{\textbf{+1.0}} \\ \midrule
LISA-7B~\citep{lai2024lisa} & 74.9 & 65.1 & 67.9 \\
w/ MosaicFusion & 76.9 & 67.9 & 69.7 \\
\emph{vs. baseline} & \textcolor{ForestGreen}{\textbf{+2.0}} & \textcolor{ForestGreen}{\textbf{+2.8}} & \textcolor{ForestGreen}{\textbf{+1.8}} \\ \midrule
GLaMM-7B~\citep{rasheed2024glamm} & 79.5 & 72.6 & 74.2 \\
w/ MosaicFusion & 80.6 & 73.6 & 75.0 \\
\emph{vs. baseline} & \textcolor{ForestGreen}{\textbf{+1.1}} & \textcolor{ForestGreen}{\textbf{+1.0}} & \textcolor{ForestGreen}{\textbf{+0.8}} \\ \bottomrule
\end{tabular}
\vspace{+10pt}
\end{table}

\noindent\textbf{Scalability in extremely large-vocabulary size.}
We further scale up MosaicFusion from the data axis to a recent more challenging V3Det dataset~\citep{wang2023v3det} to examine its effectiveness in extremely large-vocabulary size.
Specifically, V3Det is a vast vocabulary visual detection dataset containing 13204 categories, which is 10 times larger than the existing large vocabulary object detection dataset, such as 1203 categories in LVIS. It has 183k images in the training set, 29k images in the validation set, and 29k images in the testing set. For the open-vocabulary detection setting, the 13204 categories are split into 6709 base categories for training and 6495 novel categories for testing.
We follow the default setup in~\cite{wang2023v3det} to train an open-vocabulary detector on V3Det. Specifically, we use SGD optimizer with a momentum of 0.9, a weight decay of 0.0001, and a learning rate of 0.02. We train all models using the standard $2\times$ schedule with a batch size of 32. Table~\ref{tab:analysis-v3det} reports the results.
MosaicFusion significantly outperforms F-VLM baselines across different backbones (+3.5\% AP$_{\text{novel}}$ for ResNet-50, +7.5\% AP$_{\text{novel}}$ for ResNet-50x64). It is worth noting that the performance improvements on V3Det are more pronounced than LVIS, demonstrating its promising scalability in more challenging data landscapes.

\noindent\textbf{Extension to multimodal understanding tasks.}
Given the recent promise of large multimodal models (LMMs), we go beyond standard instance segmentation tasks and show that our MosaicFusion can also be potentially extended to multimodal understanding tasks.
Specifically, we take referring expression segmentation (RES) as an example, which involves segmenting objects based on free-form human language descriptions.
RES benchmarks~\citep{kazemzadeh2014referitgame,mao2016generation} include three commonly-used datasets, \ie, RefCOCO, RefCOCO+, and RefCOCOg. Compared with standard instance segmentation datasets, they have natural language expressions referring to specific objects in images. To generate the data with the corresponding format, we modify our pipeline by adding an additional referring expression generation process.
Specifically, since the location of each object is specified in each region during synthesis and known as priori, we use the following generic templates to generate referring expressions at the same time: \romannumeral1) ``$c$'', \romannumeral2) ``top/bottom left/right $c$'', \romannumeral3) ``$c$ on top/bottom left/right'', and \romannumeral4) ``the $c$ to the left/right of the $c'$ and above/below the $c''$'', where $c$ is the category name, $c'$ and $c''$ are category names in the same image with adjacent regions to $c$.
Take the synthetic image in Fig.~\ref{fig:pipeline} as an example, we can generate the following referring expressions for the ``easel'' category, respectively: \romannumeral1) ``easel'', \romannumeral2) ``top left easel'', \romannumeral3) ``easel on top left'', and \romannumeral4) ``the easel to the left of the seaplane and above the parrot''.
We generate four referring expressions for each object.

After generating both the visual and textual data, we choose several state-of-the-art LMM baselines to verify the effectiveness of our MosaicFusion. Specifically, we build our method upon F-LMM~\citep{wu2024f}, LISA~\citep{lai2024lisa}, and GLaMM~\citep{rasheed2024glamm}. We follow the default setup in each baseline and fine-tune with our data. For F-LMM, we choose DeepseekVL-1.3B~\citep{lu2024deepseek} as the frozen LMM and fine-tune a mask head for 8 epochs using a batch size of 8, with gradient clipping at a max norm of 1.0. We use AdamW optimizer with betas as (0.9, 0.999), a weight decay of 0.01, and a learning rate of 0.0001. For LISA, we fine-tune for 10 epochs using a batch size of 16, and a gradient accumulation step of 10. We use AdamW optimizer with betas as (0.9, 0.95), a weight decay of 0, and a learning rate of 0.0003. For GLaMM, we fine-tune for 5 epochs using a batch size of 16, and a gradient accumulation step of 10. We use AdamW optimizer with betas as (0.9, 0.95), a weight decay of 0, and a learning rate of 0.0003.

As shown in Table~\ref{tab:analysis-lmm}, MosaicFusion consistently outperforms each LMM baseline, demonstrating its effectiveness in multimodal understanding tasks. It should be well noted that we only use several simple templates here to generate referring expressions. We expect the performance can be further improved by leveraging large language models to rewrite and generate more diverse and detailed text descriptions, which is, however, out of the scope of this paper. We leave more explorations on multimodal understanding tasks for future work.

\begin{figure*}[t]
	\centering
	\includegraphics[width=.94\linewidth]{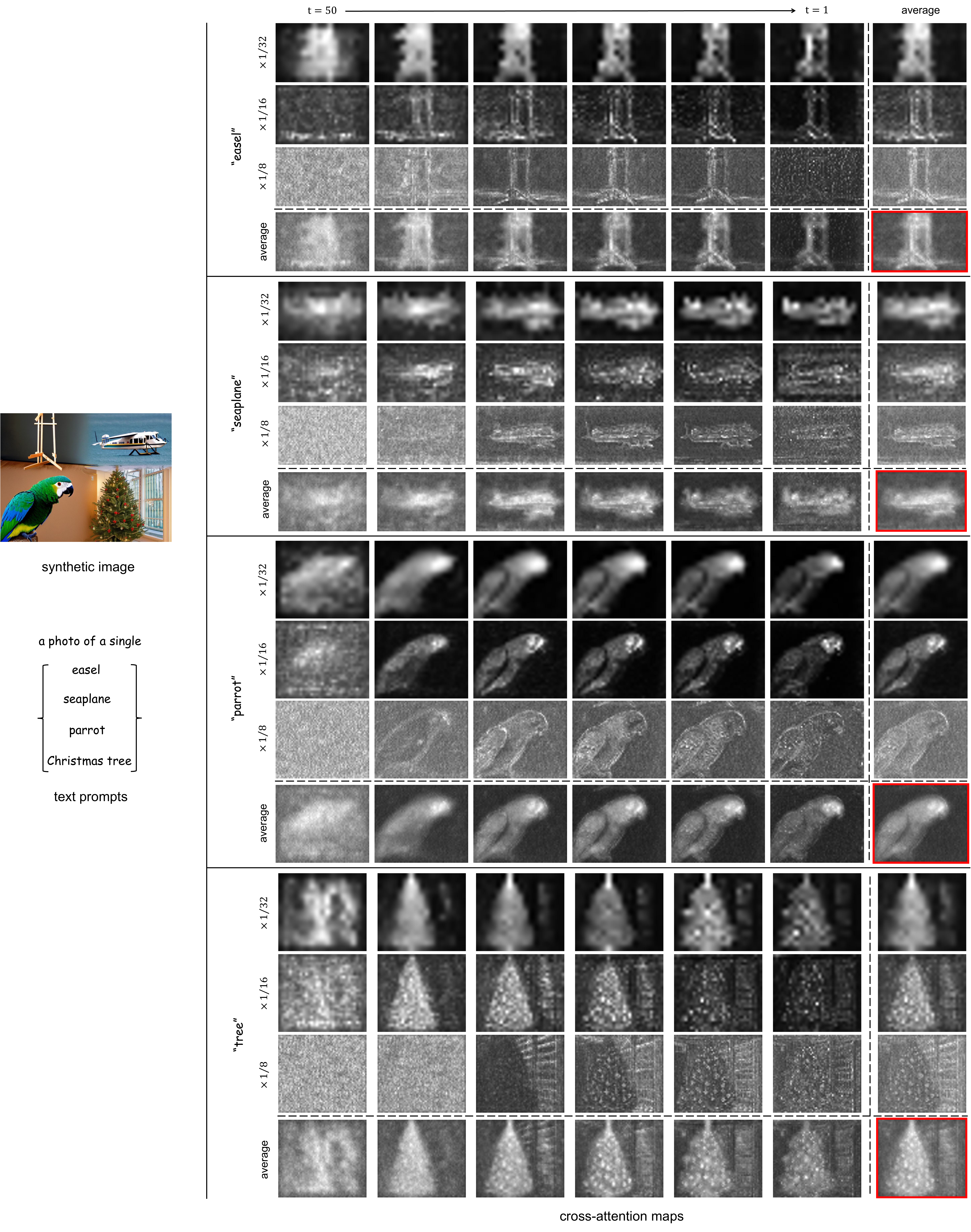}
 \vspace{-10pt}
	\caption{\textbf{Visualization of cross-attention maps with respect to each interest subject word across different time steps and layers in the diffusion process.} The time steps range from the first step $t=50$ to the last step $t=1$ in equal intervals (from \emph{left} to \emph{right}), while the layer resolutions range from $\times 1/32$ to $\times 1/8$ of the original image size (from \emph{top} to \emph{bottom}). In each entry, the \emph{last column} shows the averaged cross-attention maps across different time steps, while the \emph{last row} shows the averaged cross-attention maps across different layers. The highest-quality attention maps are produced by averaging them across both time steps and layers (\emph{bottom right} framed in \textcolor{red}{red})}
	\label{fig:supp_atten}
\end{figure*}

\begin{figure*}[t]
	\centering
	\includegraphics[width=\linewidth]{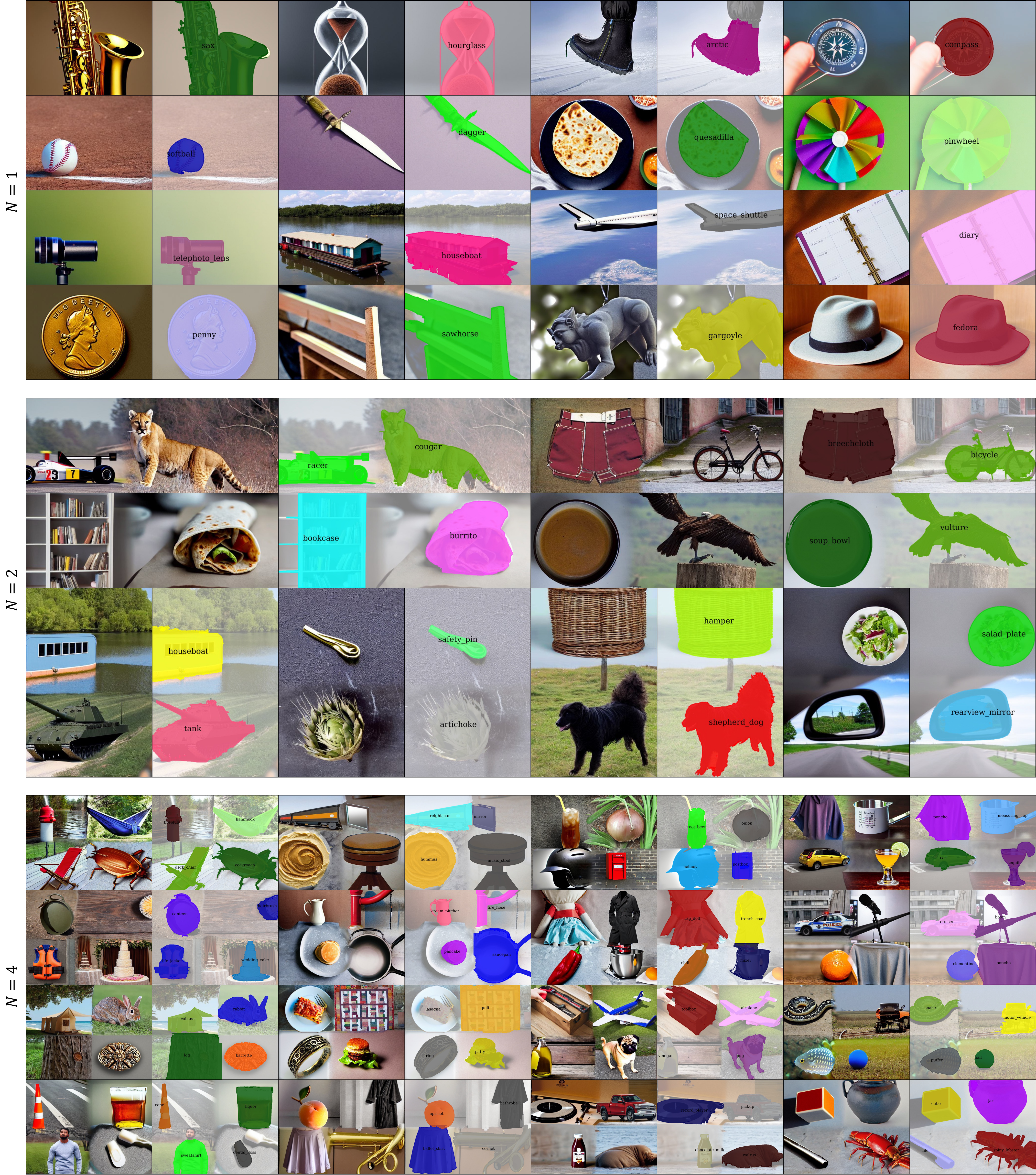}
	\caption{\textbf{Visualization of our synthesized instance segmentation dataset by MosaicFusion.} We show examples of generating $N=1,2,4$ objects per image, using the settings in Table~\ref{tab:ablation-lt-object}}
	\label{fig:supp_vlz}
\end{figure*}

\subsection{Qualitative Results}
\label{subsec:visualization}

Apart from quantitative results, we finally provide qualitative results of the cross-attention maps during the diffusion process as well as the synthesized images and masks by our MosaicFusion.
Besides, we present and analyze some failure cases during image and mask synthesis.

\noindent\textbf{Cross-attention maps.}
As illustrated in Sect.~\ref{sec:method}, cross-attention maps in the diffusion process play a key role in producing our instance segmentation masks.
We have studied the effects of attention layers and time steps on the final instance segmentation performance in Sect.~\ref{subsec:properties} (see Table~\ref{tab:ablation-lt-layer} and Table~\ref{tab:ablation-lt-time}).
In Fig.~\ref{fig:supp_atten}, we visualize some example cross-attention maps with respect to each interest word across different time steps and attention layers in the diffusion process, using the same synthetic image example as in Fig.~\ref{fig:pipeline}.
We can observe that the structure of the object is determined in the early steps of the diffusion process while more fine-grained object details emerge in the later steps. A similar phenomenon is observed across different layers. Specifically, low-resolution attention layers produce a coarse object shape while the high-resolution counterparts produce a fine-grained object edge or silhouette. In conclusion, each cross-attention map reflects the object composition to some extent. To make full use of the spatial and temporal information, we aggregate the cross-attention maps by averaging them across different time steps and layers, which produces the attention maps with the highest quality as visualized in Fig.~\ref{fig:supp_atten}. These aggregated attention maps can then serve as a good mask source to generate our final instance masks.

\noindent\textbf{Synthesized images and masks.}
Our MosaicFusion can generate multiple objects with corresponding masks in a single image.
We have studied the effect of the number of generated objects per image in Sect.~\ref{subsec:properties} (see Table~\ref{tab:ablation-lt-object}).
Here, we visualize some example results of our synthesized instance segmentation dataset by MosaicFusion.
In Fig.~\ref{fig:supp_vlz}, we show examples of generating $N=1,2,4$ objects per image, respectively. Given only interest category names, MosaicFusion can directly generate high-quality multi-object images and masks by conditioning on a specific text prompt for each 
\clearpage\noindent
region simultaneously \emph{without} any further training and extra detectors or segmentors. The synthesized instance segmentation dataset can be used to train various downstream detection and segmentation models to improve their performances, especially for rare and novel categories, as verified in Sect.~\ref{subsec:comparison}.

\begin{figure}[t]
	\centering
	\includegraphics[width=\linewidth]{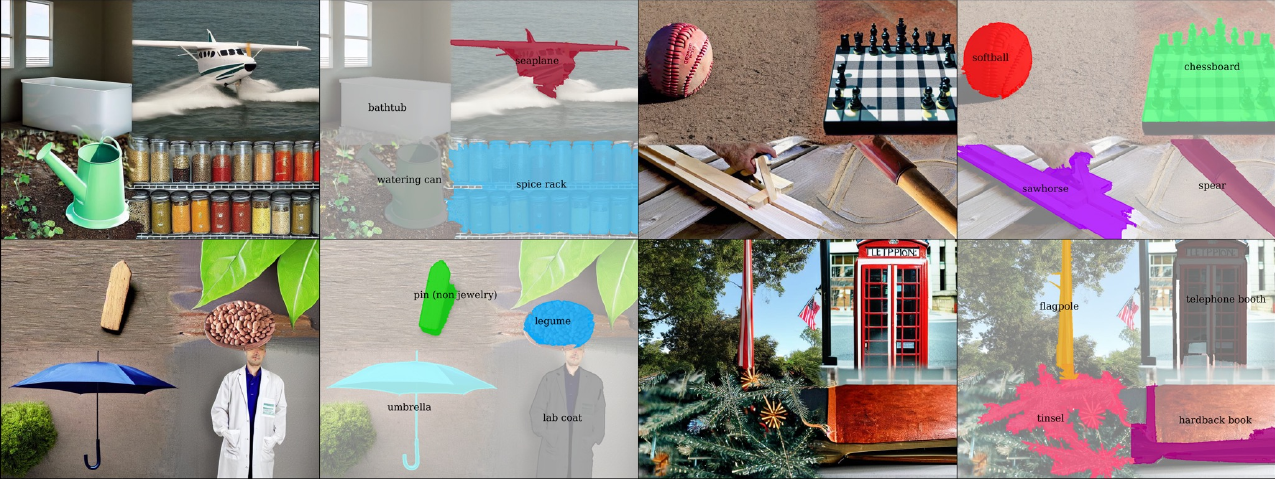}
	\caption{\textbf{Visualization of failure cases during synthesis.} MosaicFusion fails in some objects that are \romannumeral1) entangled (\eg, ``spice rack'' with spices, ``chessboard'' with chess), \romannumeral2) small (\eg, ``legume'', ``tinsel''), and \romannumeral3) abstract (\eg, ``hardback book'')}
	\label{fig:failure_case}
\end{figure}

\noindent\textbf{Failure cases.}
In Fig.~\ref{fig:failure_case}, we present some failure examples during image and mask synthesis. MosaicFusion faces challenges in dealing with some \romannumeral1) entangled (\eg, ``spice rack'' with spices, ``chessboard'' with chess), \romannumeral2) small (\eg, ``legume'', ``tinsel''), and \romannumeral3) abstract (\eg, ``hardback book'') objects. We attribute these phenomena to the inherent limitations of cross-attention in diffusion models. To mitigate these failure cases, one solution is to intervene in the generative process and refine the cross-attention maps to improve the controllability of diffusion models~\citep{hertz2023prompt,chefer2023attend,xie2023boxdiff,phung2023grounded}. Furthermore, instead of using one general prompt template, we can also leverage large language models such as GPT-4~\citep{openai2023gpt} and LLaMA~\citep{touvron2023llama,touvron2023llama2} to automatically generate more diverse and detailed text prompts to reduce the abstractness and ambiguity of some categories.

\section{Conclusion}
\label{sec:conclusion}

Instance segmentation is a fundamental task in computer vision. In this work, we have introduced a diffusion-based data augmentation method, namely MosaicFusion, for large vocabulary instance segmentation. Our method is training-free, equipped with multi-object image and mask generation capability without relying on extra models, and compatible with various downstream detection architectures. We hope our explorations can pave the way to unleash the potential of generative models for discriminative tasks.

\noindent\textbf{Limitations.}
Our study has several limitations:
1) We choose a representative text-to-image diffusion model, \ie, Stable Diffusion, due to its open-sourced implementation. 
More text-to-image diffusion models can be studied in future research.
2) We consider some representative object detection and instance segmentation models to examine the effectiveness of our method for large vocabulary instance segmentation due to resource constraints.
Essentially, MosaicFusion is orthogonal to downstream detection and segmentation models, and it can be built upon all baseline methods in Table~\ref{tab:benchmark-lt} and Table~\ref{tab:benchmark-ov} to improve their performances.
More baselines can be explored for further studies.
3) Despite the intriguing properties of MosaicFusion, the synthetic images still have an unavoidable domain gap with the real images due to the limited expressiveness of off-the-shelf diffusion models.
As the first work to generate multiple objects and masks in a single image, we leave explorations on generating more complex scene-level images with diffusion models for future work.
4) We mainly focus on synthesizing data in general scenes. It is challenging to directly apply MosaicFusion to a more specific field such as industrial or medical sectors since existing diffusion models cannot synthesize high-quality data for such domain-specific scenarios. Future attention could be devoted to generating synthetic data tailored for more specific domains.

\noindent\textbf{Data availability statements.}
All data supporting the findings of this study are available online.
The LVIS dataset can be downloaded from \url{https://www.lvisdataset.org/dataset}. 
The V3Det dataset can be downloaded from \url{https://v3det.openxlab.org.cn/}.
The RefCOCO, RefCOCO+, and RefCOCOg datasets can be downloaded from \url{https://github.com/lichengunc/refer}.
The Stable Diffusion models used to generate data are available at \url{https://github.com/CompVis/stable-diffusion} and \url{https://github.com/Stability-AI/generative-models}.

\begin{acknowledgements}
This study is supported under the RIE2020 Industry Alignment Fund – Industry Collaboration Projects (IAF-ICP) Funding Initiative, as well as cash and in-kind contribution from the industry partner(s). The project is also supported by NTU NAP and Singapore MOE AcRF Tier 2 (MOE-T2EP20120-0001, MOE-T2EP20221-0012).
\end{acknowledgements}

{\small
\bibliographystyle{spbasic}
\bibliography{bib}
}

\end{sloppypar}
\end{document}